\documentclass[10pt,conference]{IEEEtran}

\usepackage{cite}
\usepackage{graphicx}
\usepackage{subcaption}
\usepackage{xcolor}
\usepackage{booktabs}
\usepackage{enumitem}
\usepackage{soul} 
\usepackage{placeins} 
\usepackage{xurl}
\usepackage[cmex10]{amsmath}
\usepackage{verbatim}
\usepackage{dblfloatfix}

\usepackage{siunitx}
\DeclareSIUnit \voltampere { VA } 
\DeclareSIUnit \pu { pu } 

\usepackage{mathtools}
\usepackage{nicefrac}
\usepackage{array}
\usepackage{makecell}

\ifCLASSOPTIONcompsoc
 \usepackage[caption=false,font=normalsize,labelfont=sf,textfont=sf]{subfig}
\else
 \usepackage[caption=false,font=footnotesize]{subfig}
\fi
\usepackage{float}

\usepackage{fancyhdr}

\fancypagestyle{plain}{%
    \fancyhf{}
    \fancyfoot[L]{%
        \footnotesize
        \copyright 2026 IEEE. Personal use of this material is permitted. Permission from IEEE must be obtained for all other uses, in any current or future media, including reprinting/republishing this material for advertising or promotional purposes, creating new collective works, for resale or redistribution to servers or lists, or reuse of any copyrighted component of this work in other works.
    }

}

\newcommand{\subk}[2]{${#1}_{#2}$}

\newcommand\mydots{\hbox to 1em{.\hss.\hss.}}
\newcommand{\PLH}{{\mkern-2mu\times\mkern-2mu}}

\begin{document}

\title{A Memristive Synapse for Online STDP Learning and Inference in SNNs}

\author{\IEEEauthorblockN{Elia Mateu-Barriendos, Álvaro Gómez-Pau, Daniel Arumí, Rosa Rodríguez-Montañés, Salvador Manich}
\IEEEauthorblockA{Universitat Politècnica de Catalunya - BarcelonaTech (UPC)}} 

\maketitle
\thispagestyle{plain}

\begin{abstract}
This work presents a fully analog memristive synaptic circuit for online spike-timing-dependent plasticity (STDP) learning in spiking neural networks (SNNs). The proposed synapse integrates a local STDP circuit generating gradual timing-dependent conductance updates directly from pre- and post-synaptic spikes. Learning occurs during normal network operation without requiring external digital control or explicit STDP waveform synthesis.

Post-layout simulations of the memristive synapse implemented in a \qty{130}{\nano\meter} CMOS technology show spike-timing-dependent conductance adaptation during SNN operation. A 2$\times$2 SNN simulation further illustrates online neuron specialization through unsupervised learning.

\end{abstract}



\vspace{-0.4em}

\section{Introduction}
Spiking neural networks (SNNs) have emerged as a promising alternative to conventional artificial neural networks. By processing information through discrete spikes, SNNs can significantly reduce data movement and power consumption, making them particularly attractive for edge computing applications. Neuromorphic hardware aims to exploit this energy efficiency by implementing SNNs directly in silicon \cite{Indiveri2011}.

Memristive devices have attracted significant interest as artificial synapses. These two-terminal devices can exhibit analog conductance updates, resembling the adaptive behavior of biological synapses in response to neuronal activity \cite{LinaresBarranco2009}. Such gradual modulation is typically achieved using weak programming pulses, i.e., short pulses with amplitudes below those producing abrupt resistive switching transitions \cite{Brivio2022}.

Spike-timing-dependent plasticity (STDP) is a biologically inspired learning rule widely explored in memristive SNNs. In STDP, synaptic weight updates depend on the relative timing between pre- and post-synaptic spikes \cite{Bi1998}, enabling online adaptation during network operation. However, STDP-based learning generally achieves lower performance than backpropagation-based approaches for complex tasks \cite{Eshraghian2021}. Consequently, many memristive SNN implementations still rely on offline training performed on external processors \cite{Yao2020,Valentian2022}.

Previous works have experimentally demonstrated STDP \textcolor{black}{behavior} in memristive devices using overlapping bipolar voltage pulses \cite{Jo2010,Du2015a,Shooshtari2025, Ambrogio2016}, \textcolor{black}{following the approach proposed in \cite{LinaresBarranco2009}}. In this scheme, the resulting voltage across the device depends on the relative spike timing \textcolor{black}{and determines the conductance update.}  \textcolor{black}{Network-level simulations} typically use STDP models derived from experimentally measured device behavior to simulate SNNs \cite{Ambrogio2016,Querlioz2013,Guo2019,Roldan2022}. 

\begin{figure}[htbp]
\centering
\includegraphics[width=\columnwidth]{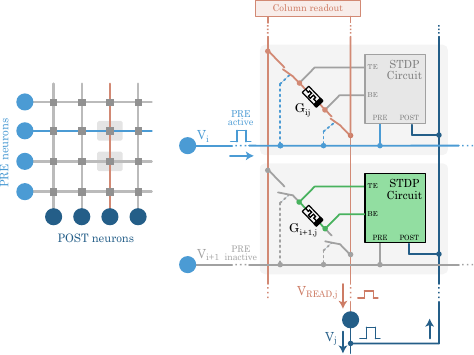}
\captionsetup{belowskip=-13pt}
\caption{SNN architecture and synaptic operation. Left: generic two-layer memristive SNN. Right: zoom-in of a column illustrating two synapses. While PRE is active (upper synapse), the memristor contributes to the column readout. Otherwise (lower synapse), the memristor is connected to the local STDP circuit.}
\label{fig:stdp-snn}
\end{figure}

\begin{figure*}[htbp]
\centering
    \begin{subfigure}{0.2\textwidth}
        \centering
        \includegraphics[width=\linewidth]{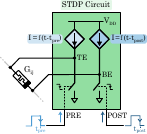}
        \caption{Conceptual representation.}
        \label{fig:scheme-a}
    \end{subfigure}
    \hfill
    \begin{subfigure}{0.39\textwidth}
        \centering
        \includegraphics[width=\linewidth]{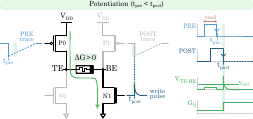}
        \caption{Circuit-level operation for potentiation.}
        \label{fig:scheme-b}
    \end{subfigure}    
    \hfill
    \begin{subfigure}{0.39\textwidth}
        \centering
        \includegraphics[width=\linewidth]{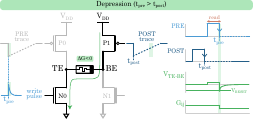}
        \caption{Circuit-level operation for depression.}
        \label{fig:scheme-c}
    \end{subfigure}     
    \captionsetup{belowskip=-8pt}
    \caption{Operation of the proposed STDP circuit.}
    \label{fig:stdp-concept}
\end{figure*}

However, network-level simulations using STDP models do not fully capture hardware non-idealities or how inference and learning operations coexist within the same synaptic circuit. Moreover, gradual conductance modulation remains challenging in practice due to device variability, abrupt switching, and asymmetric SET/RESET characteristics \cite{Brivio2022}.

\textcolor{black}{Several works have proposed STDP circuits for memristive synapses, including experimental results with an external memristor emulator \cite{Wu2015} and circuit-level simulations \cite{Chakraborty2023a,Acciarito2017}. However, \cite{Chakraborty2023a,Wu2015} require multiple voltage references, digital control and separate inference and learning phases, increasing circuit complexity and limiting fully local and online learning. In \cite{Pedretti2017}, STDP learning is experimentally demonstrated in a $16\PLH1$ SNN implemented on a PCB, but the timing-dependent update is induced by synthetic waveforms generated by a microcontroller, without a  \mbox{dedicated STDP circuit.}}

This work proposes a synaptic circuit for online STDP learning in memristive SNNs. The circuit is fully analog, local and event-driven, generating timing-dependent conductance updates directly from pre- and post-synaptic spikes without digital control or STDP waveform synthesis. Post-layout simulations in \qty{130}{\nano\metre} CMOS demonstrate gradual potentiation and depression during SNN operation. A 2$\times$2 SNN simulation further illustrates online unsupervised specialization.


\section{The synaptic circuit}

\subsection{Synapse architecture}

The proposed synapse consists of a memristor, a local STDP circuit, and two switches. As illustrated in Fig.~\ref{fig:stdp-snn}, the switches are controlled by the pre-synaptic signal (PRE) and selectively connect the memristor either to the column readout circuitry or to the STDP circuit.

While PRE is active, the synapse contributes to inference and the STDP circuit remains electrically decoupled from the memristor (upper synapse in Fig.~\ref{fig:stdp-snn}). This work assumes the voltage-sensing scheme from \cite{mem10}, although the STDP operation is independent of the readout method. Each post-synaptic neuron receives a voltage \subk{V}{READ,j} encoding the weighted sum $\sum_i X_i G_{ij}$, and the resulting POST spikes are fed back to the corresponding column STDP circuits.

When PRE is inactive, the memristor is connected to the STDP circuit (lower synapse in Fig.~\ref{fig:stdp-snn}). As described in the next subsection, conductance updates can only occur at the falling edges of PRE and POST, so learning does not interfere with the read operation.

\subsection{STDP operation}

The operating principle of the STDP circuit is illustrated in Fig.~\ref{fig:stdp-concept}. A conceptual representation is first introduced in Fig.~\ref{fig:scheme-a}. The circuit consists of two symmetric branches driven by PRE and POST, respectively, connected to the top (TE) and bottom (BE) electrodes of the memristor. Conceptually, each branch contains a time-dependent current source and a switch.

At the circuit level, PMOS transistors P0/P1 implement the current sources, controlled by analog synaptic traces that decay over time and define the potentiation and depression time windows (Fig.~\ref{fig:scheme-b}-c). NMOS transistors N0/N1 implement the switches, enabled by short write pulses generated at the falling edges of PRE and POST ($t_{pre}$ and $t_{post}$, respectively).

At the falling edge of PRE (POST), the corresponding trace is activated and a write pulse is generated in the PRE (POST) branch. If the pulse overlaps with an active trace in the \textit{complementary} branch, current flows through the memristor. The magnitude of the resulting voltage drop depends on the trace amplitude and the memristor state, while its polarity depends on the trace-pulse combination. If the voltage exceeds the SET or RESET threshold, potentiation (PRE trace + POST pulse) or depression (POST trace + PRE pulse) occurs. 

Thus, the conductance update depends on both the memristor state and the relative timing between PRE/POST spikes:

\vspace{-0.4em}

\begin{equation}
    \Delta G_{ij} = f(G_{ij}, t_{post}-t_{pre})
\end{equation}

\vspace{-0.4em}

Conversely, if the complementary trace is inactive, no current flows through the memristor. Therefore, isolated PRE or POST spikes produce no conductance update.

Fig.~\ref{fig:scheme-b} illustrates potentiation \mbox{($t_{pre}<t_{post}$)}, where a POST write pulse overlaps with an active PRE trace. Although the POST trace \textcolor{black}{also} initiates at $t_{post}$, current through P1 does not significantly contribute to \textcolor{black}{programming} because N1 provides the lowest-resistance path. Conversely, Fig.~\ref{fig:scheme-c} shows depression \mbox{($t_{pre}>t_{post}$)}, where a PRE write pulse overlaps with an active POST trace. In both cases, read occurs while PRE is active and does not interfere with conductance modulation.

\begin{figure}[htbp]
\centering
\includegraphics[width=\columnwidth]{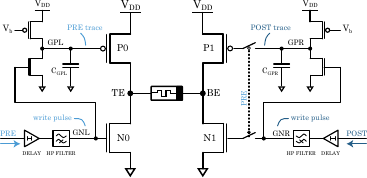}
\caption{High-level STDP circuit schematic.}
\label{fig:stdp-schematic}
\end{figure}

Fig.~\ref{fig:stdp-schematic} shows a simplified STDP circuit schematic. The write pulses are generated from PRE (POST) signals through high-pass filters, and simultaneously (1) activate the corresponding NMOS switch N0 (N1) and (2) initiate the PRE (POST)-synaptic trace by rapidly discharging capacitor $C_{GPL}$ ($C_{GPR}$). The capacitor then progressively recharges toward \subk{V}{DD}, generating the time-dependent gate voltage that controls P0 (P1).

To avoid perturbations during read operation, P1/N1 in the POST branch are temporarily disabled while PRE is active, preventing unintended current flow through the memristor due to a POST spike arriving during PRE. Importantly, the write-pulse generation and trace circuitry remain active so POST events occurring during read are still captured. Additional delay before the high-pass filters further prevents overlap between read and learning operations.


\section{Results}
The proposed synaptic circuit was implemented in a \qty{130}{\nano\metre} CMOS technology \cite{ihp} and submitted for fabrication. It comprises the memristor, the STDP circuit and the two synaptic switches. A \textcolor{black}{voltage-based readout circuit} \cite{mem10} was included to validate the read operation. 

\begin{figure}[htbp]
\centering
\includegraphics[width=0.95\columnwidth]{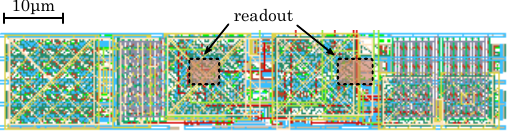}
\captionsetup{belowskip=-6pt}
\caption{Layout of the implemented circuit (\qty{1218.6}{\micro\metre\squared}).} 
\label{fig:stdp-layout-read}
\end{figure}

Fig.~\ref{fig:stdp-layout-read} shows the circuit layout, with an area of \qty{1175.1}{\micro\metre\squared} excluding the readout circuit. The implemented values for the STDP circuit are: \mbox{$C_{GPL}=C_{GPR}=\qty{250}{\femto\farad}$} (synaptic trace capacitors); \mbox{$C=\qty{100}{\femto\farad}$, $R=\qty{400}{\kilo\ohm}$} (high-pass filters); $(W/L)_{P0}=6/0.3$, $(W/L)_{N0}=12/0.3$, $(W/L)_{P1}=12/0.3$, and $(W/L)_{N1}=4.4/0.3$ (transistor dimensions given as $W/L$ in \si{\micro\meter}/\si{\micro\meter}). The two switches are implemented as CMOS transmission gates. 

\vspace{-0.6em}

\textcolor{black}{Energy consumption was evaluated at the schematic-level. Static power consumption is around \qty{138}{\micro\watt}, while the energy per conductance update is in the order of \qty{64}{\pico\joule}, measured over the interval between $t_{pre}$ and $t_{post}$ causing the update, with read energy excluded. The energy remains relatively consistent across different conductances and relative timings.}

\textcolor{black}{Table~\ref{tab:comparison} compares the proposed circuit with other STDP circuits for memristive synapses. The proposed circuit has comparable area and energy consumption, although only an order-of-magnitude comparison is possible due to differences in the reported energy calculations. This work further provides post-layout simulations including the memristor and enables local and online STDP learning.}

In this work, the memristor is simulated using the Verilog-A model from the PDK and calibrated by IHP from static measurements. However, it might not fully reproduce the device dynamics under short pulses, potentially leading to differences in the experimentally observed STDP curve while preserving the underlying conductance modulation mechanism.

\begin{table}[htbp]
\centering
\caption{\textcolor{black}{Comparison with other reported STDP circuits.}}
\label{tab:comparison}
\setlength{\tabcolsep}{3pt}
\renewcommand{\arraystretch}{0.7}
\resizebox{\columnwidth}{!}{%
\begin{tabular}{lccc m{3.3cm}}
\toprule
\textbf{Work} & \textbf{CMOS / Area [$\mu$m$^2$]} &
\textbf{Power} & \textbf{Energy} & \textbf{STDP results} \\
\midrule

\cite{Chakraborty2023a} &
\makecell{\qty{65}{\nano\meter} / \qty{1436}{\micro\meter\squared} \\
(\qty{5744}{\micro\meter\squared} @ \qty{130}{\nano\meter})} &
\qty{118}{\micro\watt} & \qty{6.4}{\pico\joule} &
\mbox{STDP characteristic } (schematic-level) \\
\midrule
\cite{Wu2015} &
\qty{180}{\nano\meter} / -- &
-- & \qty{9.3}{\pico\joule} &
\mbox{STDP behavior in a 2$\times$1} \mbox{SNN (experimental with} memristor emulator) \\
\midrule
\cite{Acciarito2017} &
\makecell{\qty{90}{\nano\meter} / \qty{480}{\micro\meter\squared} \\
(\qty{1001}{\micro\meter\squared} @ \qty{130}{\nano\meter})} &
\qty{540}{\micro\watt} & -- &
\mbox{STDP behavior in a 2$\times$1} SNN (schematic-level) \\

\midrule

\makecell{\textbf{This} \\ \textbf{work}} &
\makecell{\qty{130}{\nano\meter} / \qty{1175}{\micro\meter\squared}} &
\textbf{\qty{138}{\micro\watt}} & \qty{64}{\pico\joule} &
\mbox{STDP characteristic (post-} \mbox{layout); STDP behavior in a} 2$\times$2 SNN (schematic-level) \\

\midrule
\end{tabular}%
}
\vspace{-1em} 
\end{table}

\subsection{Post-layout STDP characterization}

\begin{figure}[htbp]
\centering
\includegraphics[width=\columnwidth]{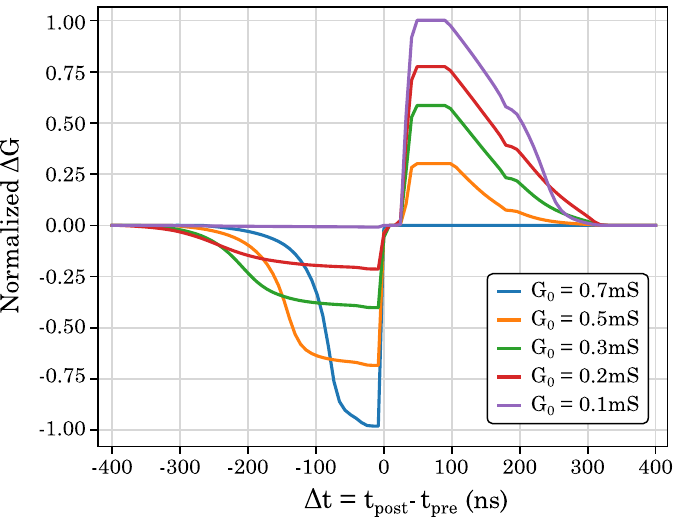}
\captionsetup{belowskip=-9pt}
\caption{STDP characteristic showing normalized conductance change $(G - G_0)/(G_{\max} - G_{\min})$ as a function of relative spike timing and initial conductance. Each curve corresponds to the cumulative effect of 50 identical PRE--POST spike pairs with fixed relative timing $\Delta t$.} 
\label{fig:stdp-curve}
\end{figure}

Fig.~\ref{fig:stdp-curve} shows the STDP characteristic for different initial memristor conductance states within the operating range of \qtyrange[range-phrase=~--~]{0.1}{0.7}{\milli\siemens}. Positive and negative spike timing differences produce potentiation and depression, respectively, demonstrating the expected STDP behavior. The effective learning window is approximately \qty{400}{\nano\second}. The conductance update magnitude depends on both the relative spike timing and the initial memristor state, following the asymmetric SET/RESET behavior of the device. 

\begin{figure}[htbp]
\centering
\includegraphics[width=\columnwidth]{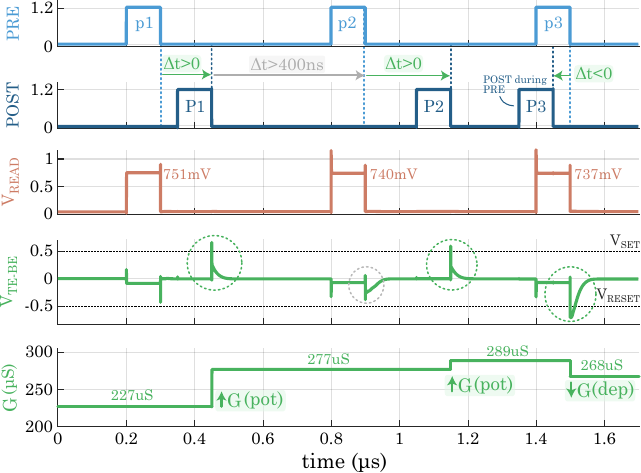}
\captionsetup{belowskip=-9pt}
\caption{Post-layout transient simulation of the proposed synaptic circuit for different relative spike timings.}
\label{fig:stdp-sim}
\end{figure}

Fig.~\ref{fig:stdp-sim} presents post-layout transient simulations for different PRE/POST timing relationships, with an initial conductance $G_0=\qty{227}{\micro\siemens}$. While PRE is active, the memristor contributes to the column read voltage $V_{READ}$. STDP conductance updates occur at the falling edges of PRE and POST according to their relative timing.

In Fig.~\ref{fig:stdp-sim}, the first pair (p1--P1) produces potentiation since P1 occurs within the active trace window generated by p1. The second pair (P1--p2) lies outside the learning window, producing insufficient \subk{V}{TE-BE} to exceed \subk{V}{RESET} and hence no update. The larger timing difference in the third potentiation event (p2--P2) results in a smaller conductance increase. Finally, P3--p3 demonstrates depression while showing that POST spikes during PRE do not perturb the read operation.

\subsection{Online learning in a 2$\times$2 SNN}
\begin{figure}[htbp]
\centering
\includegraphics[width=\columnwidth]{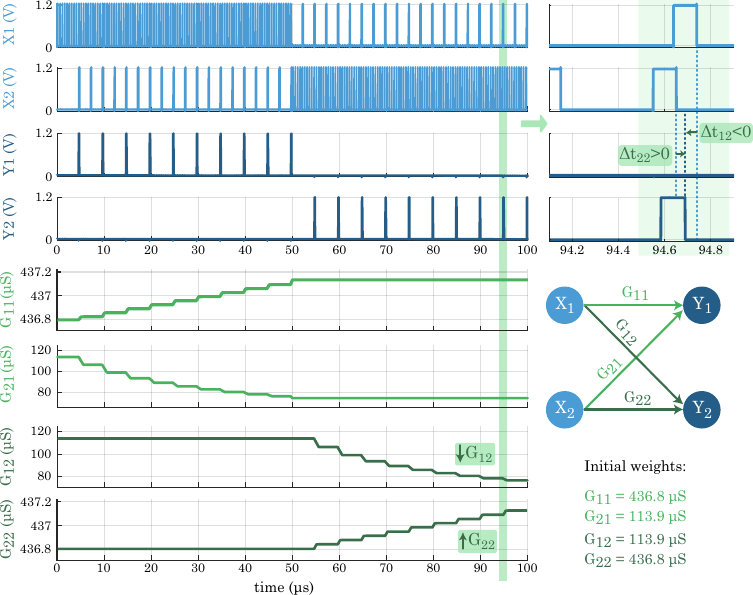}
\captionsetup{belowskip=-9pt}
\caption{Simulation of a 2$\times$2 memristive SNN demonstrating online unsupervised weight adaptation through local STDP.}
\label{fig:snn-sim}
\end{figure}

A schematic simulation of a 2$\times$2 memristive SNN including the proposed synaptic circuit was performed (Fig.~\ref{fig:snn-sim}). Synaptic weights $G_{11}$ and $G_{22}$ were initially set higher to establish a preference for the corresponding input--neuron pairs. The neurons were implemented as integrate-and-fire units with lateral inhibition.

Complementary spike patterns were applied to $X_1$ and $X_2$ during two consecutive phases, producing potentiation or depression through local STDP. Consequently, $G_{11}$ and $G_{22}$ were reinforced, while $G_{12}$ and $G_{21}$ were weakened, leading to selective neuron responses.

\section{Conclusion}
This work presented a memristive synaptic circuit enabling fully analog, event-driven online STDP learning in SNNs. The proposed architecture generates timing-dependent conductance updates directly from pre- and post-synaptic signals, avoiding external digital control and explicit STDP waveform synthesis. Switches controlled by the pre-synaptic signal selectively connect the memristor to the readout path or to the local STDP circuit, enabling online learning without perturbing inference.

The proposed synaptic circuit was implemented in \qty{130}{\nano\metre} CMOS. Post-layout simulations demonstrated gradual timing-dependent potentiation and depression with effective learning windows of approximately \qty{400}{\nano\second}. A 2$\times$2 SNN \textcolor{black}{schematic-level} simulation illustrated neuron specialization through online unsupervised STDP. Future work will focus on experimental characterization of the fabricated prototype and \textcolor{black}{validation of STDP learning in larger-scale SNNs}.


\section*{Acknowledgment}
This work has been supported by PID2022-141391OB-C22 funded by MCIN/AEI/10.13039/501100011033/FEDER, UE. The corresponding author gratefully acknowledges the Universitat Politècnica de Catalunya and Banco Santander for the financial support of her predoctoral FPI-UPC grant.

\vspace{-0.1em}

\bibliographystyle{IEEEtran}
\bibliography{papers_overleaf, additional_bib}

\vfill

\end{document}